\documentclass[10pt,journal,compsoc]{IEEEtran}
\ifCLASSOPTIONcompsoc
  \usepackage[nocompress]{cite}
\else
  \usepackage{cite}
\fi
\usepackage[pdftex]{graphicx}
\DeclareGraphicsExtensions{.pdf,.jpg}
\ifCLASSINFOpdf
\else
\fi
\usepackage{color}
\usepackage{threeparttable}
\usepackage{amsmath}
\usepackage{amsfonts,amssymb} 
\usepackage{stmaryrd}

\begin{document}
%
\title{VoxSegNet: Volumetric CNNs for Semantic Part Segmentation of 3D Shapes}
%
%
%
%

\author{Zongji~Wang,
	    Feng~Lu,~\IEEEmembership{Member,~IEEE}
\thanks{
Correspondence should be addressed to Feng Lu.
\hfil\break
The authors are with the State Key Laboratory of Virtual Reality Technology and Systems, School of Computer Science and Engineering, Beihang University. Feng Lu is also with the International Research Institute for Multidisciplinary Science, Beihang University, Beijing 100191, China.
E-mail: \{wzjgintoki, lufeng\}@buaa.edu.cn}
}

\markboth{Journal of \LaTeX\ Class Files,~Vol.~14, No.~8, August~2015}%
{Shell \MakeLowercase{\textit{et al.}}: Bare Demo of IEEEtran.cls for Computer Society Journals}

\IEEEtitleabstractindextext{
\begin{abstract}
   Voxel is an important format to represent geometric data, which has been widely used for 3D deep learning in shape analysis due to its generalization ability and regular data format. However, fine-grained tasks like part segmentation require detailed structural information, which increases voxel resolution and thus causes other issues such as the exhaustion of computational resources. In this paper, we propose a novel volumetric convolutional neural network, which could extract discriminative features encoding detailed information from voxelized 3D data under a limited resolution. To this purpose, a spatial dense extraction (SDE) module is designed to preserve the spatial resolution during the feature extraction procedure, alleviating the loss of detail caused by sub-sampling operations such as max-pooling. An attention feature aggregation (AFA) module is also introduced to adaptively select informative features from different abstraction scales, leading to segmentation with both semantic consistency and high accuracy of details. Experiment results on the large-scale dataset demonstrate the effectiveness of our method in 3D shape part segmentation.
\end{abstract}  

}

\maketitle

\section{Introduction}

In the past few decades, with the advances in user-friendly 3D modeling tools (e.g., SketchUp) and low-cost 3D shape capturing devices, the amount of available 3D shapes on the Internet has increased significantly. This has induced much research interest in 3D shape analyzing and understanding, for which 3D shape semantic segmentation is considered as a fundamental yet challenging task. 3D shape segmentation can be defined as the partition of a 3D shape into meaningful parts, which could greatly benefit a large number of applications such as modeling \cite{Model-interchange07,Fu2017-Pose,Xu2012-FDS}, shape editing \cite{Fu2016-Edit,Yang2013-Deform} and object classification \cite{Huber2004-POC}.

The quickly increasing 3D models on the Internet also provide the opportunity to introduce deep learning methods into the field of 3D shape analysis. Inspiring by the success of convolutional neural networks (CNNs) on various computer vision tasks such as image classification \cite{Krizhevsky2012-AlexNet}, semantic segmentation \cite{LongSD14_FCN}, and image caption generation \cite{Kelvin2015-ShowTell}, an intuitive idea for 3D deep learning is to apply CNNs to extract high-quality 3D geometric features. However, directly applying CNNs to 3D shapes faces obstacles, one of which is the irregular data formats of 3D shapes.

In order to solve the problem due to irregular 3D formats, several different representations for 3D shapes are investigated. 
\textit{Volumetric representation} depicts a shape's occupancy in a gridded cubic space, thus 3D convolution could be applied to a voxelized shapes \cite{maturana_iros_2015,Zhirong15CVPR}. 
\textit{Multi-view representation} models a 3D shape by a set of 2D images rendered from different viewpoints, which could be fed into 2D CNNs \cite{Kalogerakis2017-ShapePFCN, Bai2016GIFTAR, Su2015-MCN}. 
For \textit{triangular mesh representation}, there are methods extending the CNNs to the graphs defined by the mesh \cite{Boscaini2015-LCD,Boscaini2016-LSC,Masci2015-GCN}. Such convolution is conducted in a non-Euclidean space, extracting features robust to isometric deformation.
Without 3D convolution, there are also methods directly processing \textit{unorganized point sets} in 3D. In PointNet \cite{qi2016pointnet}, all 3D points are processed individually sharing the same set of network weights, and then global operation such as max pooling is applied to extract global features.

\textbf{Why volumetric.} Volumetric data naturally encode the spatial distribution of 3D shapes. Similar to the pixels in a bitmap, voxels excel at representing regularly sampled spaces, explicitly describing the spatial relationships between elements forming a shape. This makes it convenient to apply 3D convolution on voxels. In addition, volumetric data could be transformed from other 3D data formats through an effortless sampling procedure, which demonstrates its generalization ability. These advantages make volumetric representation a good choice for 3D deep learning.

Besides the volumetric representation, meshes and point clouds are popular 3D data formats. However, mesh convolution methods require smooth manifold meshes as input, which is not so common in today's large shape collections. Point clouds lack neighborhood structure, making it hard to extract contextual information from. Although multi-view representation agrees with the human's habit of observing 3D shapes through visual cues, the viewpoint selection and lighting parameters influence the system's robustness.
Therefore, in this paper, we select volumetric representation to model the 3D shapes.

Nowadays, volumetric CNNs have already gained success in shape classification and retrieval \cite{Zhirong15CVPR, maturana_iros_2015, qi2016volumetric}.
Although detailed structures have been lost during the convolution procedure, the features encoding global information are already sufficient for tasks like classification. However, for part segmentation, it is desired to preserve as much detailed information as possible, motivating recent papers \cite{Riegler2017, Wang-2017-OCNN, Graham18} to work on high resolution volumetric data. To reduce the expensive computational and memory cost caused by larger voxel resolution (e.g., 64 or larger), they have to use either more complex data formats such as octree or carefully designed convolutional operation. But there is still inevitable loss of detailed information due to the sub-sampling during feature extraction.

In this paper, we propose a novel deep network architecture (VoxSegNet) for volumetric semantic segmentation. The key idea of our method is to extract discriminative features encoding detailed information under a limited resolution. To this purpose, we propose two network modules.
First, a Spatial Dense Extraction (SDE) module is designed to extract discriminative features from sparse volumetric data. Without reducing the spatial resolution, this module effectively alleviates the loss of detailed structural information caused by sub-sampling operations in CNNs.
Second, we introduce an Attention Feature Aggregation (AFA) module, which uses attention mechanism to aggregate features according to their different levels of abstraction. This module enables us to contextually select more informative components from an input signal. Integrating the above modules, our method could obtain dense prediction with both semantic consistency and high accuracy of details.

The contributions of this paper include:
1) A novel approach for volumetric object semantic segmentation which integrates the Spatial Dense Extraction and the Attention Feature Aggregation.
2) A volumetric feature extraction method which could preserve detailed structural information.
3) An attention-based feature aggregation method to adaptively select informative features from different abstraction levels.
4) Experiments on the large-scale dataset demonstrate the effectiveness of our method in 3D shape part segmentation.

\section{Related Work}
3D shape part segmentation has long been studied. Many early researches for mesh segmentation have a major concern finding a single effective feature for mesh label identification \cite{Katz2003,BenChen2008,HWAG2009,Shapira2010,Zhang2012,Au2012}. However, the meshes in different 3D models may vary remarkably. A single feature is often insufficient to cover all kinds of scenarios. To address this problem, Kalogerakis et al. \cite{Kalogerakis2010} presented a learning-based method to segment and label 3D meshes by combining various geometric features. Later, Guo et al. \cite{Guo2015} applied deep CNNs to a group of hand-engineered geometric features organized into a 2D matrix. These methods rely on human-designed geometric features extracted from high quality smooth manifold meshes. In this section, we mainly cover works of shape part segmentation using deep learning methods. 

Recently, with the availability of large-scale 3D shape collections, there have been increasing research interest in solving 3D shape analyzing tasks via deep learning methods. Volumetric representation of 3D shapes makes it convenient to automatically extract deep features from raw data.
Wu et al. \cite{Zhirong15CVPR} trained a convolutional deep belief network on voxelized shapes for object classification, shape completion and next best view prediction. 
Maturana et al. \cite{maturana_iros_2015} and Qi et al. \cite{qi2016volumetric} proposed volumetric CNN architectures to learn discriminative features to identify or classify shapes. These methods restricted their input voxel data to a limited grid size like ${30^3}$ or ${32^3}$, due to the high computational and memory cost.
Li et al. \cite{Li2016} proposed a field probing scheme to simultaneously train the weights and locations of the filters, reducing the computational complexity.
Graham et al. \cite{Graham15, Graham18} utilized the 3D sparse CNNs that apply CNN operations to active voxels.
Riegler et al. \cite{Riegler2017} presented a non-uniform volumetric representation utilizing the concept of octree structure, making it possible to compute 3D CNNs with high-resolution inputs.
Recently, Wang et al. \cite{Wang-2017-OCNN} presented a 3D CNN based on octree representation, which can largely improve the computation efficiency.

Although volumetric CNNs have been widely used in tasks like shape classification and retrieval, it is relatively less prefered in semantic segmentation. Partially because the limited spatial resolution may set an upper bound to the non-trivial fine-grained and dense prediction task. With the sparsity of volumetric data being studied, it becomes possible to compute high-resolution 3D CNNs more efficiently. Several works paid attention to 3D CNNs for semantic segmentation.
Riegler et al. \cite{Riegler2017} performed 3D semantic segmentation on a colored 3D point cloud facades dataset. The point clouds were mapped to grid-octree structure and then fed into a U-shaped encoder-decoder convolutional neural network to predict the label for each voxel. Wang et al. \cite{Wang-2017-OCNN} also used a U-shaped convolutional neural network for shape part segmentation, achieving the state-of-the-art results with the ${64^3}$ resolution of leaf octant. In \cite{Graham18}, utilizing submanifold sparse convolution, the authors trained FCN \cite{LongSD14_FCN} and U-Net \cite{RonnebergerFB15_Unet} for part semantic segmentation of voxelized volumes. While inputs with high-resolution benefit part segmentation task, there would be inevitable loss of detailed information due to the sub-sampling operation during feature extraction.

\begin{figure*}[!t]
	\centering
	\includegraphics[width=1.0\linewidth]{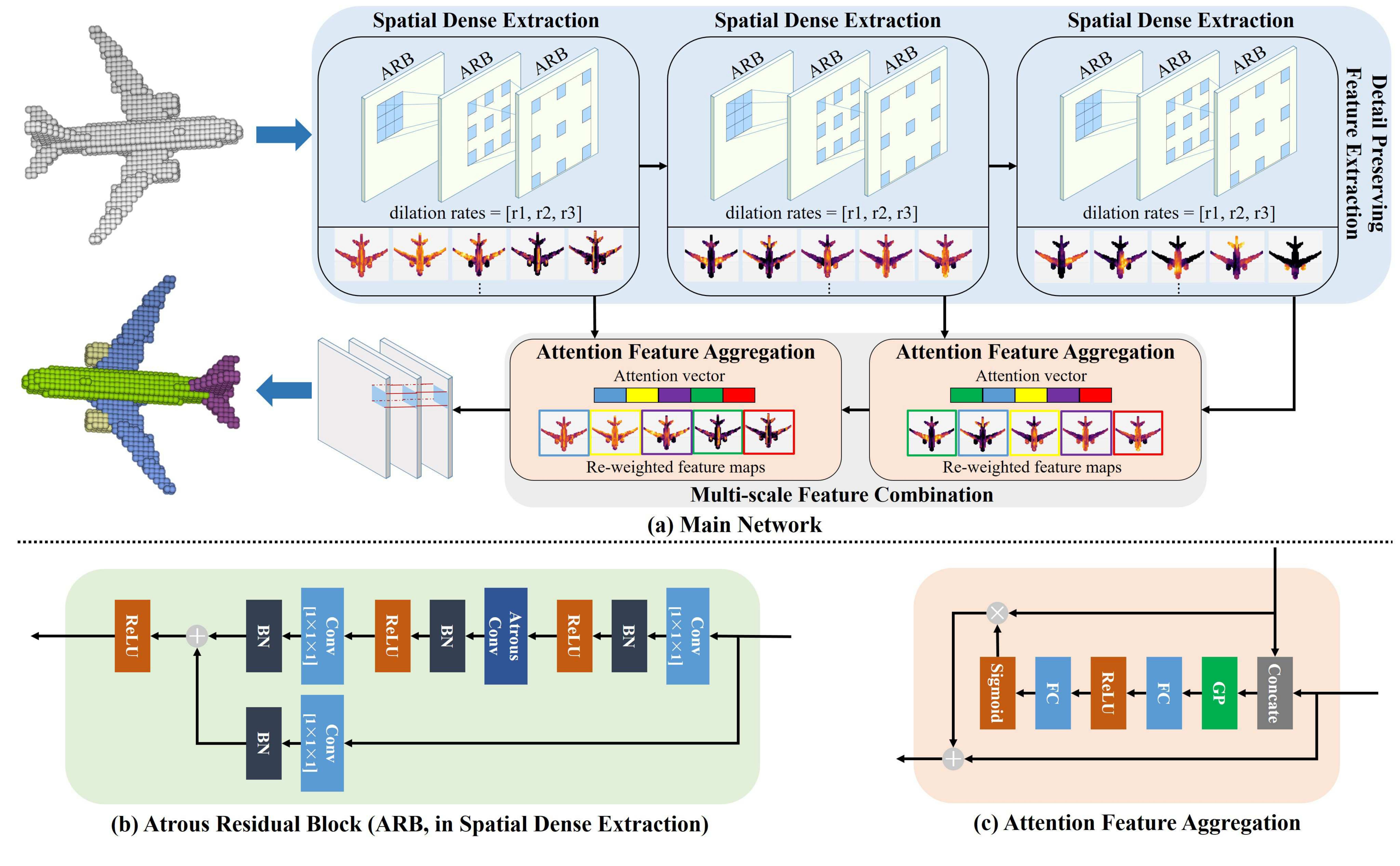}
	\caption{An overview of the proposed Voxel Segmentation Network (VoxSegNet). (a) depicts the main network architecture. The input voxelized shape is passed through the Spatial Dense Extraction units to extract multi-scale features preserving detailed information. Some of the visualized filter responses are shown within the units. After that, the features from different extraction stages are combined by the Attention Feature Aggregation units. Finally, three $1\times1\times1$ convolutional layers are used to predict part labels for each voxel. In (b), the inner structure of the Atrous Residual Block (ARB) is presented. ARB is the basic component of the Spatial Dense Extraction unit. (c) is the inner structure of the Attention Feature Aggregation unit. This unit utilizes high stage filters to guide the feature selection of low stage.}
	\label{network}
\end{figure*}

Besides the volumetric CNNs, there are methods segmenting point clouds without voxelizing the input. PointNet \cite{qi2016pointnet} operated on unordered points set directly. The independently processed points are aggeregated into global feature by max-pooling. In the following work PointNet++ \cite{qi2017pointnetpp}, the authors improved PointNet by incorporating local dependencies and hierarchical feature learning in the network. Kd-Networks \cite{KlokovL17} built a Kd-tree by recursively partitioning the space along the axis of the largest variation on the input point clouds. Huang et al. \cite{Huang18_RSN} proposed a Recurrent Slice Network (RSNet) to directly segment point clouds. The sliced points were input to a stack of bidirectional RNNs sequentially, generating features by interacting with neighboring points.

\section{Method}

%
%
%
%

\subsection{Problem statement}
A voxelized shape can be easily obtained from a point cloud or a triangular mesh, capturing the spatial occupancy within a 3D space regularized to 3D lattices. Denote $V_{i,j,k}$ as the state value on the specific discrete voxel coordinate $(i,j,k)$, reflecting the state whether this grid is occupied. This is an intuitive and effective representation for a 3D object, without suffering from the variance introduced by different meshing methods or the lack of topology introduced by point clouds representation.

Given a 3D object represented by volumetric matrix $V$, the goal of part segmentation is to assign a part category label to each occupied element $\{v\in V|V_{location(v)}=1\}$. In this paper, we design a deep learning framework to model the function $f(v)=l$, where $f:V\mapsto L$, $l\in L$ and $L=\{1,2,\dots,K\}$ is the set of part labels.

\subsection{Network Architecture}
Our full network architecture is visualized in Figure \ref{network}. Taking the voxelized shape as input, the Spatial Dense Extraction (SDE) units are used to extract discriminative features encoding detailed information from raw data. As shown in Figure \ref{network} (a), an SDE unit consists of stacked Atrous Residual Blocks (ARBs) with user specified dilation rates for each ARB. The inner structure of an ARB is depicted in Figure \ref{network} (b). The $1\times1\times1$ convolutional layers are used to change the channel number of feature maps, reducing the computational complexity of the atrous convolutional layer. Batch normalization and ReLU activation are used after the first $1\times1\times1$ convolutional layer as well as the atrous convolutional layer. The sequence of SDEs extract multi-scale features, which encode information from low-level geometry to high-level semantics and preserve the spatial resolution of input signals. Such features from different extraction stages are combined by the Attention Feature Aggregation (AFA) units. In an AFA unit, after concatenation between feature maps from high- and low- stages, global pooling layer is applied to abstract features channel-wise. Then, fully connected layers are used to compute the attention weights for low-stage features. Finally, stacked $1\times1\times1$ convolutional layers are used to predict the semantic label per voxel, which could be considered as Multilayer Perceptrons (MLPs). Using Softmax as the cost function, the optimization goal can be represented as $\min \limits_{\theta}{J(\theta)}$:
\begin{equation}
\centering
J(\theta)=-\displaystyle\frac{1}{m}\sum\limits_{i=1}^{m} \sum\limits_{j=1}^{K} 1\{y^{(i)}=j\}log\frac{e^{\hat{f}_j(v^{(i)}|\theta)}}{\sum_{l=1}^{K}{e^{\hat{f}_l(v^{(i)}|\theta)}}},
\label{softmax}
\end{equation}
where $y^{(i)}$ is the ground truth label of voxel $i$, $\hat{f}_j(\cdot)$ is the last layer output in channel $j$, and $m$ is the voxel number of the input 3D shape.

Our network has two key modules: the Spatial Dense Extraction (SDE) and the Attention Feature Aggregation (AFA). In the rest of this section, we detailedly introduce the characteristics of the proposed modules.

\subsection{Atrous Convolution in 3D}
\label{3D_Atrous}
Convolutional Neural Networks (CNNs) have shown promising performance in both $2D$ and $3D$ semantic segmentation. However, the previous methods \cite{LongSD14_FCN,RonnebergerFB15_Unet,Wang-2017-OCNN,Graham18} usually use CNNs in a fully convolutional fashion, which introduces spatial pooling and strided convolution repeatedly, thus, would reduce the spatial resolution of the extracted feature maps. The semantic segmentation is a dense prediction task in which a high-resolution output is desired. In order to recover the spatial resolution from the feature maps, upsampling methods like upconvolution and unpooling are proposed. But the predicted results recovered from low-resolution feature maps still suffer from problems such as blur and edge misalignment.

Atrous (dilated) convolution is originally introduced in \textit{algorithme \`atrous}, which is an algorithm for wavelet decomposition. Recently, it is widely applied in 2D segmentation tasks to extract semantic-rich feature maps, instead of using stacked spatial pooling and strided convolution \cite{ChenPSA17_RethinkAtrous, YuKoltun2016}. Atrous convolution could enlarge the receptive field by inserting `holes' in the convolutional kernels, thus is able to avoid down-sampling operations and benefit from fewer training kernel factors (sparse kernel). We would like to take advantage of atrous convolution in 3D shape analysis.

In 3D, atrous convolution is defined as:
\begin{equation}
\centering
g[i,j,k]=\sum_{l=1}^{L}\sum_{m=1}^{M}\sum_{n=1}^{N}{f[i+rl,j+rm,k+rn]h[l,m,n]},
\label{3D_dilated}
\end{equation}
where $f[i,j,k]$ is the input signal, $g[i,j,k]$ is the output signal, $h[l,m,n]$ denotes the filter of size $L\times M \times N$, and $r\in \mathbb{Z}_+$ corresponds to the dilation rate used to sample $f[i,j,k]$. In a deep CNN, the feature maps in lower layers contain more spatial structural information due to the small filed-of-view, while the ones in higher layers contain more high-level semantic information \cite{lecun2015deep}. By changing the dilation rate, we cloud control the filed-of-view of corresponding convolutional kernel easily. 

\begin{figure}[!t]
	\centering
	\includegraphics[width=1.0\linewidth]{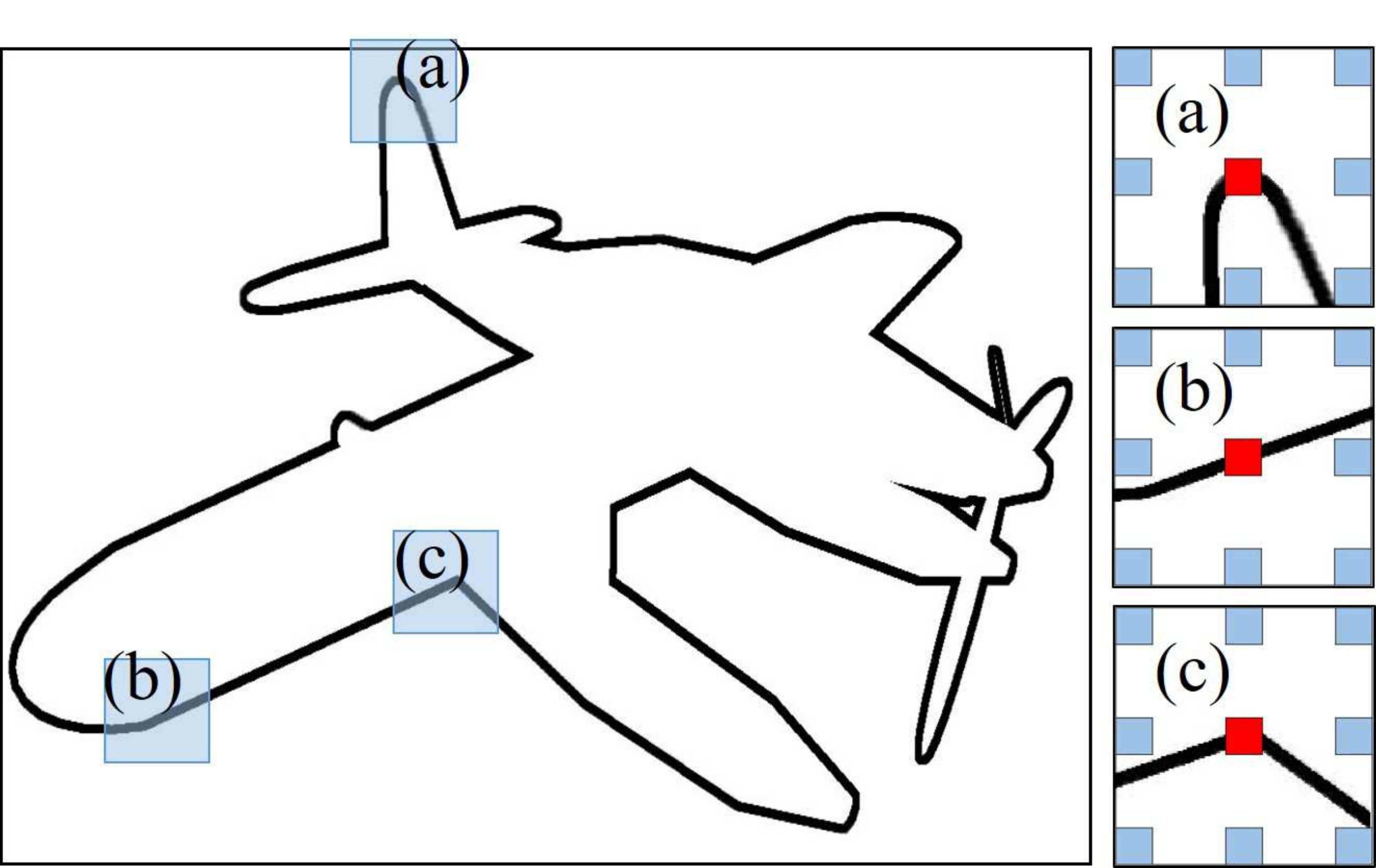}
	\caption{Atrous convolution may face problem extracting discriminative features from sparse data. Without loss of generality, we take a 2D case as an example. An atrous kernel with dilated rate 3 is used to extract features from an airplane boundary image. At three locations (a), (b), and (c), only the weight at the center of the kernel is activated, resulting in the same convolution value. This means the filter is failed to capture different local patterns from the sparse boundary image.}
	\label{sparsity}
\end{figure}

The traditional convolutional network is usually applied to data like photos and videos, which frequently comprise densely populated grids. However, voxelized volumes of a 3D object often have sparse structure, since the corresponding surface mesh representation can be modeled as 2D manifolds in a 3D Euclidean space. Notice that the main idea of dilated convolution is to insert `holes'(zeros) between pixels (or voxels) in convolutional kernels to enlarge the receptive fileds. The atrous convolutional kernel would only capture the information with non-zero weights. With the dilation rate increasing, the non-zero weights in the kernel become further in location. While applying to data with sparse spatial structure, it is highly possible an atrous kernel could not cover any local information due to \textit{the sparsity of both the input signal and the kernel weights}. In Figure \ref{sparsity}, a $3 \times 3$ 2D atrous convolutional kernel with dilation rate 3 is used to extract features from the 2D boundary of an airplane (boundary image has sparse structure). As we can see, the filter is failed to capture the discriminative features from different local patterns $(a)$, $(b)$ and $(c)$. Specifically, only the center of the kernel is activated at each of these locations, thus degenerating the atrous convolution to $1\times1$ convolution. This effect could decrease the robustness and discrimination of the extracted features. A detailed discussion could be found in Section \ref{ablation}.

\subsection{Spatial Dense Extraction}
Inspired by \cite{wang2017understanding}, we introduce the Spatial Dense Extraction (SDE) unit to alleviate information loss caused by sparse kernels. In detail, an SDE unit consists of multiple stacked atrous convolutional layers with dilation rates of $[r_1, r_2, ..., r_n]$ and kernel size $K \times K \times K$. The final size of the receptive field of an SDE should fully cover a cubic region without any holes.

To model the sparsity of an atrous convolutional kernel, the ``maximum distance between active weights'' is defined as
\begin{equation}
\begin{split}
M_l &= max[|M_{l+1}-2r_l|, r_l], \\
M_n &= r_n,
\end{split}
\label{dist_hole}
\end{equation}
in which $|\cdot|$ returns 1-norm value, and $l\in\{1,2,...,n\}$ represents the layer index in the SDE. In 1-D situation, the nearest active weights from a higher layer's atrous kernel define a line segment. And current layer's active weights split this line segment into at most two kinds of segments, depending on their lengths: the distance between current layer's active weights $r_l$, and that between active weights from current layer and higher layer $|M_{l+1}-2r_l|$.

In order to get full coverage fields in the top layer of an SDE, $r$ and $M$ should subject to several conditions:
\begin{equation}
\begin{split}
1\leq r_1\leq r_2 &\leq \cdots \leq r_n, \\
M_2 &\leq K.
\end{split}
\label{condition}
\end{equation}
With $M_2$ not greater than the kernel size (active notes number along an axis), for an extreme condition, the holes would always be filled by the lowest layer convolution with dilation rate $r_1=1$ in an SDE.

Different from 2D images which contain dense and smooth signals, 3D volumetric shapes are sparse, making it harder to extract discriminative features using sparse kernels. Using stacked atrous kernels with dilated rates subjecting to equation \ref{condition}, the features could be ensured to encode discriminative information from the sparse 3D data.
In our network architecture, the spatial dense extraction is the basic component to extract different stages of features from the input voxelized volumes. In one stage, we use three residual blocks \cite{Kaiming2016}, each of them is a bottleneck structure with an atrous convolutional layer with dilation rate $r_l$. For example, we can build an SDE unit with dilation rates $r=[1,2,3]$, in which there are three residual blocks stacked together with dilation rate $[r_1=1,r_2=2,r_3=3]$ respectively. We use multiple SDE units to extract different stages of multi-scale features.


\subsection{Attention Feature Aggregation}

In convolutional neural networks, features from deep layers learn more about high-level semantic information, while features of shallow layers keep rich spatial structural details. To accurately segment an object into semantic parts, it is necessary to combine multi-level features together.
However, previous encoder-decoder networks for fine-grained and dense tasks usually directly integrate multi-level features indiscriminately (e.g., U-Net \cite{RonnebergerFB15_Unet}). The equal weights for different channels of integrated features are defective due to the redundant details and distractions from different parts. 

\begin{figure}[!t]
	\centering
	\includegraphics[width=1.0\linewidth]{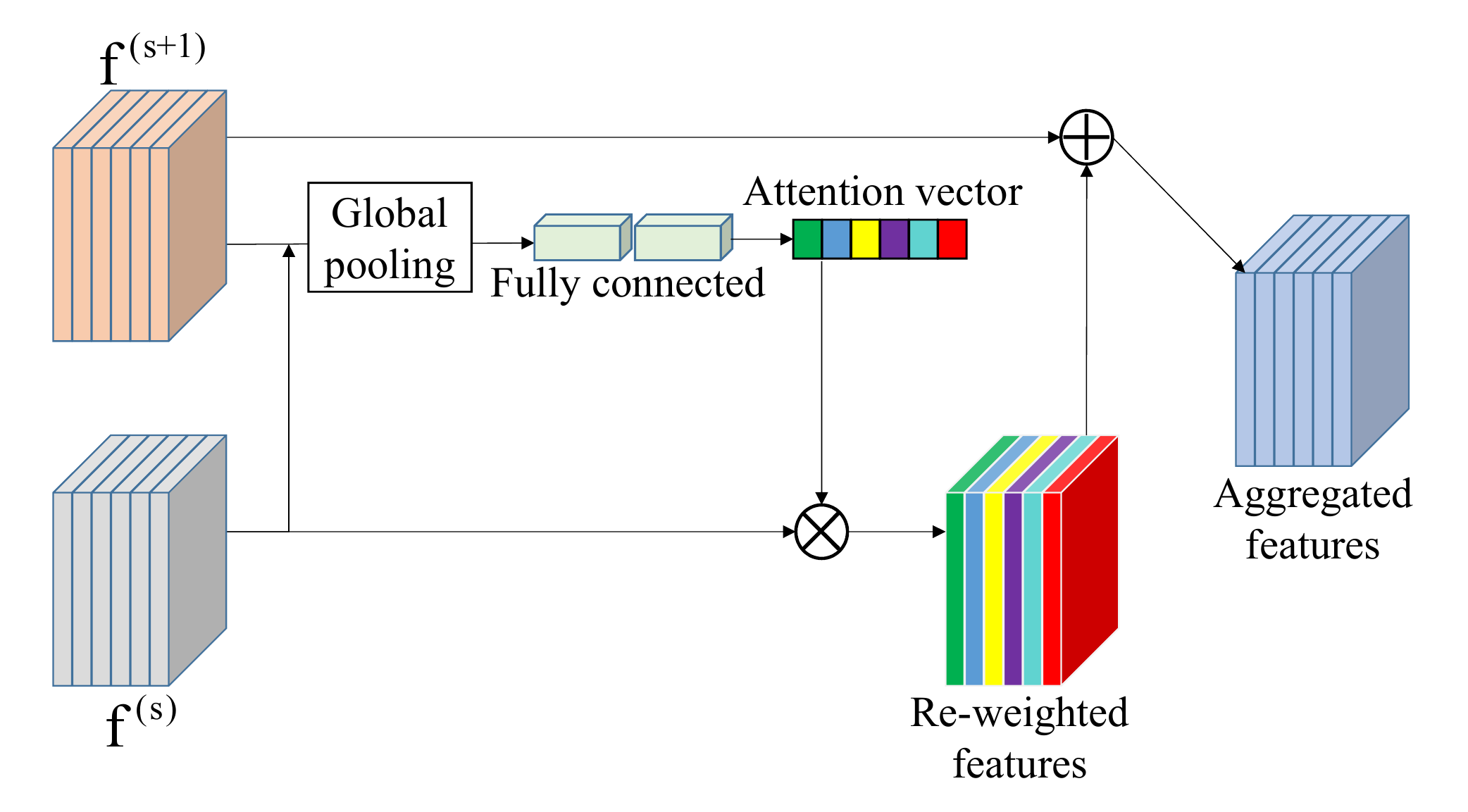}
	\caption{Architecture of attention feature aggregation. $\mathbf{f}^{(s)}$ and $\mathbf{f}^{(s+1)}$ are feature maps from two adjacent SDE stages. The low-stage feature maps $\mathbf{f}^{(s)}$ are re-weighted by attention vector, and then added to the high-stage ones, generating more informative aggregated features.}
	\label{AFA}
\end{figure}

To address the problem, we propose the Attention Feature Aggregation (AFA) unit leveraging attention mechanism \cite{itti2001computational, hu2017squeeze}, which could extract the informative features and suppress the indiscriminative ones. 
As shown in Figure \ref{AFA},
the AFA unit takes both low- and high-stage feature maps as input to compute the channel attention weights. Then, the low-stage feature maps are re-weighted by the channel attention weights and added to the high-stage ones. Using the high-level semantic information to guide the selection of low-level detailed information, multi-scale features with stronger discriminative ability are aggregated, leading to a segmentation which is both semantically consistent and accurate. 

To formulate the attention feature aggregation, we unfold convolutional features $\mathbf{f}$ as $\mathbf{f}=[\mathbf{f}_1,\mathbf{f}_2,...,\mathbf{f}_C]$, where $\mathbf{f}_i\in \mathbb{R}^{W\times H\times D}$ is the $i_{th}$ slice of $\mathbf{f}$ and $C$ is the channel number. First, the input feature maps from two feature extraction stages are concatenated and then passed through a global average pooling layer to extract the channel-wise global context:
\begin{equation}
\mathbf{z} = \frac{1}{W\times H\times D}\llbracket \mathbf{f}^{(s)},\mathbf{f}^{(s+1)}\rrbracket ,
\end{equation}
in which $\llbracket\cdot \rrbracket$ means concatenation, $\mathbf{f}^{(s)}$ is the feature maps from the $s_{th}$ extraction stage, and $\mathbf{z}\in \mathbb{R}^{C_s+C_{s+1}}$ is the channel-wise global feature vector. Then, two fully connected layers are exploited to learn the aggregated feature of each channel:
\begin{equation}
\begin{split}
\mathbf{u}_1 &= \mathbf{W}_1\bullet\mathbf{z}+\mathbf{b}_1, \\
\mathbf{u}_2 &= \mathbf{W}_2\bullet ReLU(\mathbf{u}_1)+\mathbf{b}_2,
\end{split}
\end{equation}
where $\bullet$ denotes matrix multiplication, $\mathbf{W}_1\in\mathbb{R}^{C_s\times (C_{s}+C_{s+1})}$ and $\mathbf{W}_2\in\mathbb{R}^{C_s\times C_{s}}$ are fully connected weights, $\mathbf{b}_1, \mathbf{b}_2\in\mathbb{R}^{C_s}$ are the bias parameters. To define the attention for the channels of feature maps, a Sigmoid operation is applied to $\mathbf{u}_2$ to generate attention of each channel $i$:
\begin{equation}
\mathbf{a}(i) = \frac{e^{\mathbf{u}_2(i)}}{e^{\mathbf{u}_2(i)}+1},
\end{equation}
where $\mathbf{u}_2(i)$ is the feature of channel $i$ and $\mathbf{a}\in\mathbb{R}^{C_s}$ is the channel-wise attention vector.

In the decoding phase of our method, attention feature aggregation units are applied progressively between adjacent spatial dense extraction (SDE) units. The attentive information from one unit serves as a guidance for the next to adaptively generate new attentions. Finally, multi-level features are aggregated to form a more discriminative ones:
\begin{equation}
\mathbf{\tilde{f}}^{(s)} = \mathbf{f}^{(s)}\bullet\mathbf{a}\oplus \mathbf{f}^{(s+1)},
\end{equation}
in which $\oplus$ represents element-wise addition operation. In the prediction phase, unit stride convolutional layers followed by a Softmax operation are used to generate the voxel-wise part label prediction.

\section{Experiments and Discussion}

\begin{figure*}[!t]
	\centering
	\includegraphics[width=1.0\linewidth]{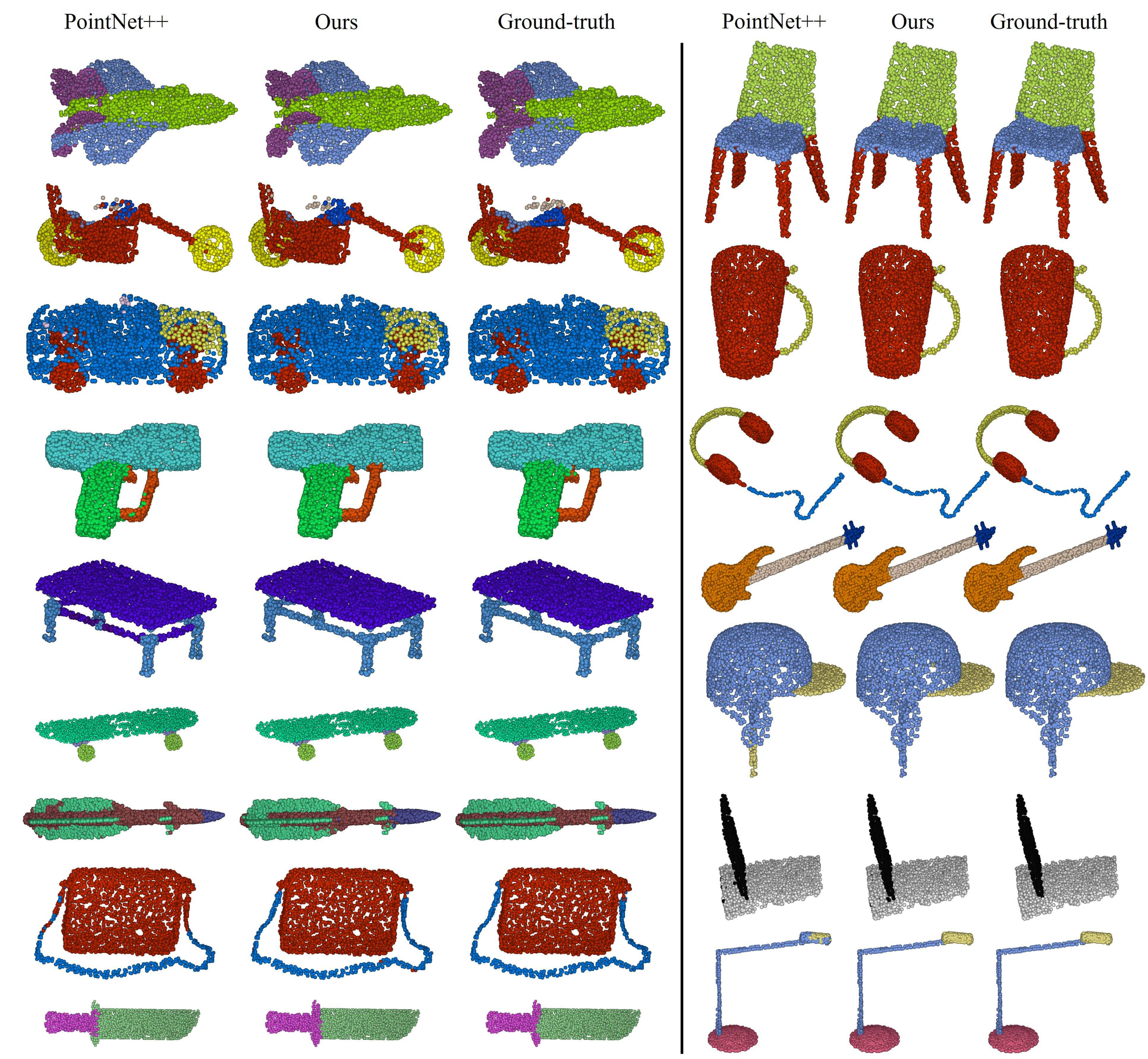}
	\caption{Qualitative comparison between PointNet++ and our method (VoxSegNet) on object-part segmentation.}
	\label{qualitative_eval}
\end{figure*}

\textbf{Dataset.} We conduct an experiment on a large-scale shape part annotation dataset introduced by Yi et al.\cite{Yi16}, which augments a subset of the ShapeNet models with semantic part annotations. The dataset contains $16$ categories of shapes, with 2 to 6 parts per category. In total there are $16,881$ models and $50$ parts. Wang et al. \cite{Wang-2017-OCNN} augmented the dataset by projecting the point label back to the triangle faces of the corresponding 3D mesh, and condense the point cloud by uniformly re-sampling the triangle faces. We use the same augmented dense point clouds data to generate voxel data, and use the same training/test split with \cite{Yi16}.

\noindent
\textbf{Training details.} In a data preprocessing step, each point cloud is centered and rescaled to a unit sphere. After that, the normalized point cloud is voxelized into a 3D grid with a user-specified size. In this paper, we use voxel size 48. The dilation rates of the three SDE units in our VoxSegNet are set to be $[1,1,1]$, $[1,3,5]$, and $[1,3,5]$ respectively. We train the network using Adaptive Moment Estimation (Adam) optimization \cite{KingmaB14_Adam} with batch size 4. The initial learning rate is $0.001$, $\beta_1$ is $0.9$, $\beta_2$ is $0.999$, and $\epsilon$ is $10^{-8}$. We augment the training data by rotating each shape $n\pi/6$ around the upright axis, where $n\in\{0,1,2,\dots,11\}$.

\subsection{Comparison to state-of-the-arts}
To evaluate the part segmentation quality, the predicted results of voxelized volumes are projected to the corresponding point clouds.
We then compute the Intersection over Union (IoU) as the metric. Specifically, for each shape, all the part class IoUs are averaged to obtain the object IoU. Per category average IoU is computed by averaging across all shapes with the certain category label. Then an overall average IoU is computed through a weighted average of per category IoU. The weights are just the number of shapes in each category. 

The numerical comparison is conducted with a learning-based technique \cite{Yi16} which uses per-point local geometric features and correspondences between shapes, and five recent deep learning based methods \cite{yi2016syncspeccnn,qi2016pointnet,qi2017pointnetpp,Graham18,Wang-2017-OCNN}. As shown in Tabel \ref{tabel_1}. Our proposed method performs better or comparable to other methods in most of the categories. Specifically, we achieve the best overall average IoU. And in individual categories, we rank the best in 9 out of 16 categories, the second best in 3 categories, and the third in the rest 4 categories. We achieve promising object part predictions without post-processing such as Conditional Random Field (CRF) refinement as is done in O-CNN \cite{Wang-2017-OCNN}.

Figure \ref{qualitative_eval} presents some examples of segmentation results of our VoxSegNet compared with those of PointNet++ \cite{qi2017pointnetpp}. As we can see, our results are visually better in most cases. Specifically, the boundary between different parts can be separated more accurately. For example, the lamp base, head, and the connection part are predicted properly by our method, while the PointNet++ result shows some artifacts around the lamp head (see the right-bottom of Figure \ref{qualitative_eval}). In addition, our VoxSegNet can separate out the fin (near the rocket nose) from the frame of the rocket while PointNet++ cannot. Similar observations can be made for other shapes such as the motorbike, pistol, and bag, etc.

\begin{table*}[!t]
	\renewcommand{\arraystretch}{1.2}
	\caption{Object part segmentation results, measured by averaged IoU (\%).}
	\label{tabel_1}
	\begin{threeparttable}
		\resizebox{\textwidth}{20mm}{
			\begin{tabular}{c|c|c c c c c c c c c c c c c c c c}
				\hline
				\  &\textbf{mean} &plane &bag &cap &car &chair &e.ph. &guitar &knife &lamp &laptop &motor &mug &pistol &rocket &skate &table \\
				\hline
				\# shapes &- &2690 &76 &55 &898 &3758 &69 &787 &392 &1547 &451 &202 &184 &283 &66 &152 &5271 \\
				\hline
				Yi2016\cite{Yi16}   &81.4 &81.0 &78.4 &77.7 &75.7 &87.6 &61.9 &\underline{92.0} &85.4 &82.5 &95.7 &70.6 &91.9 &\textbf{85.9} &53.1 &69.8 &75.3 \\
				SpecCNN\cite{yi2016syncspeccnn}  &84.7 &81.6 &81.7 &81.9 &75.2 &90.2 &74.9 &\textbf{93.0} &86.1 &\underline{84.7} &95.6 &66.7 &92.7 &81.6 &\underline{60.6} &\textbf{82.9} &82.1 \\
				PointNet\cite{qi2016pointnet} &83.7 &83.4 &78.7 &82.5 &74.9 &89.6 &73.0 &91.5 &85.9 &80.8 &95.3 &65.2 &93.0 &81.2 &57.9 &72.8 &80.6 \\
				PointNet++\cite{qi2017pointnetpp} &85.1 &82.4 &79.0 &\underline{87.7} &77.3 &90.8 &71.8 &91.0 &85.9 &83.7 &95.3 &71.6 &94.1 &81.3 &58.7 &76.4 &82.6 \\
				SSCN\cite{Graham18}     &\underline{86.0} &84.1 &83.0 &84.0 &\textbf{80.8} &\underline{91.4} &\underline{78.2} &91.6 &\textbf{89.1} &\textbf{85.0} &\underline{95.8} &\underline{73.7} &95.2 &\underline{84.0} &58.5 &76.0 &82.7 \\
				O-CNN\cite{Wang-2017-OCNN} &85.9 &\underline{85.5} &\underline{87.1} &84.7 &77.0 &91.1 &\textbf{85.1} &91.9 &\underline{87.4} &83.3 &95.4 &56.9 &\underline{96.2} &81.6 &53.5 &74.1 &\underline{84.4} \\
				\hline
				Ours     &\textbf{87.5} &\textbf{86.2} &\textbf{88.7} &\textbf{91.9} &\underline{79.8} &\textbf{92.0} &76.5 &\underline{92.0} &86.4 &84.2 &\textbf{96.1} &\textbf{78.4} &\textbf{96.3} &83.7 &\textbf{65.4} &\underline{77.0} &\textbf{86.2} \\
				\hline
			\end{tabular}
		}
		
		\begin{tablenotes}
			\footnotesize
			\item[] (\textbf{Bold number}: the highest score; \underline{underlined number}: the second highest score)
		\end{tablenotes}
	\end{threeparttable}
	
\end{table*}

\begin{figure*}[t]
	\centering
	\includegraphics[width=1.0\linewidth]{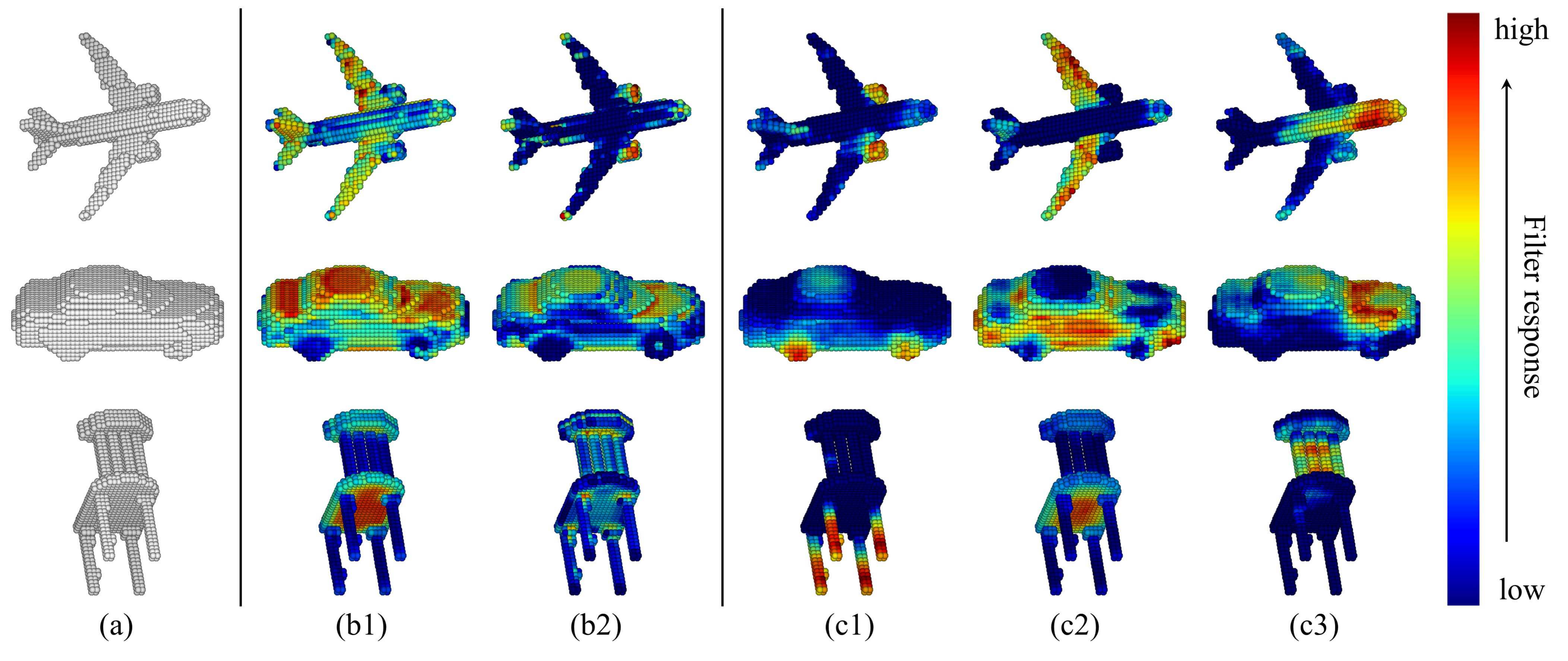}
	\caption{
	Visualization of filter responses from different feature extraction stages.
	For each row, (a) is the input voxelized shape; (b1) and (b2) show filter responses from the first SDE module; 
	(c1), (c2), and (c3) show filter responses from the second SDE module in feature extraction procedure of the proposed segmentation network.
	Note that low-stage filters capture geometric patterns while high-stage ones capture regions with semantic meanings, which could demonstrate the effectiveness of our proposed feature extraction module.}
	\label{fea_visual}
\end{figure*}

\subsection{Visualization and analysis}
\label{cnn_vis}
For image understanding using deep convolutional neural networks, it is known that the learned filters are activated when important image features appear. And the filters in different convolutional stages capture different levels of features. In other words, filters in the first convolutional layer are usually activated by the object edges in the image, while the higher level layer's filters capture more complex patterns \cite{Zeiler14_visual}. A similar phenomenon can be observed on our VoxSegNet, and it helps us to intuitively understand the segmentation network.

In Figure \ref{fea_visual}, we visualize some filters in different feature extraction stages of VoxSegNet by color-coding the responses to the input voxelized shapes. In detail, (a) is the input voxelized shape. For the first feature extraction stage, (b1) and (b2) are the responses of two filters which capture low-level geometric patterns. While the responses of filters from the second feature extraction stage (c1), (c2), and (c3) show the filters' ability to capture high-level shape features. For example, in the first row of Figure \ref{fea_visual}, filter (b1) tends to capture large planar regions such as wings of an airplane, filter (b2) tends to capture sharp and pointed areas. In stage two, filters tend to capture regions possessing semantic information, discriminating wings (c2) from fuselage (c3) and engines (c1). Similar observations can be made for the other two samples. Note that for the chair in the third row, despite that the leg and the back both consists of bars with similar low-level geometric features, (c1) and (c3) discriminate them successfully.

\subsection{Ablation study}
\label{ablation}

U-Net is a popular network structure for encoder-decoder. After the encoding phase, a U-Net uses upsampling methods such as deconvolution and unpooling to obtain feature maps with higher spatial resolution. During decoding phase, by concatenating the feature maps with the corresponding ones from the encoding phase through skip connections, the information loss caused by large convolution strides and pooling operation could be alleviated to some extent. In this section, we use U-Net as a baseline method. And the performances of several network structures are compared in order to investigate the effect of the proposed modules.

In detail, we experiment on the following network architectures:


1) U-Net 3DCNN. We use the plain encoder-decoder architecture as a baseline. In the encoding phase, 3D convolution with kernel size 3 followed by a batch normalization layer and a Rectified Linear Unit (ReLU) activation layer are used to extract features. And then a max-pooling layer with stride 2 is applied to the feature maps, reducing the spatial size to a half in order to increase the receptive field. The encoding phase contains three such $(conv, bn, relu, pooling)$ blocks. In the decoding phase, we apply 3D deconvolution (transpose convolution) to the encoded feature maps with kernel size 3 and stride 2. Batch normalization and ReLU activation are also conducted after deconvolution. The feature maps are concatenated with those from the encoding phase with the same spatial size, as the input to the next deconvolutional layer. The decoding phase consists of three $(deconv, bn, relu)$ blocks. After that, three stacked 3D convolutional layers with unit kernel size and unit stride are applied to get the voxel-wise prediction. This architecture achieves good object part segmentation results with average IoU of $83.97\%$.

\begin{figure}[!t]
	\centering
	\includegraphics[width=1.0\linewidth]{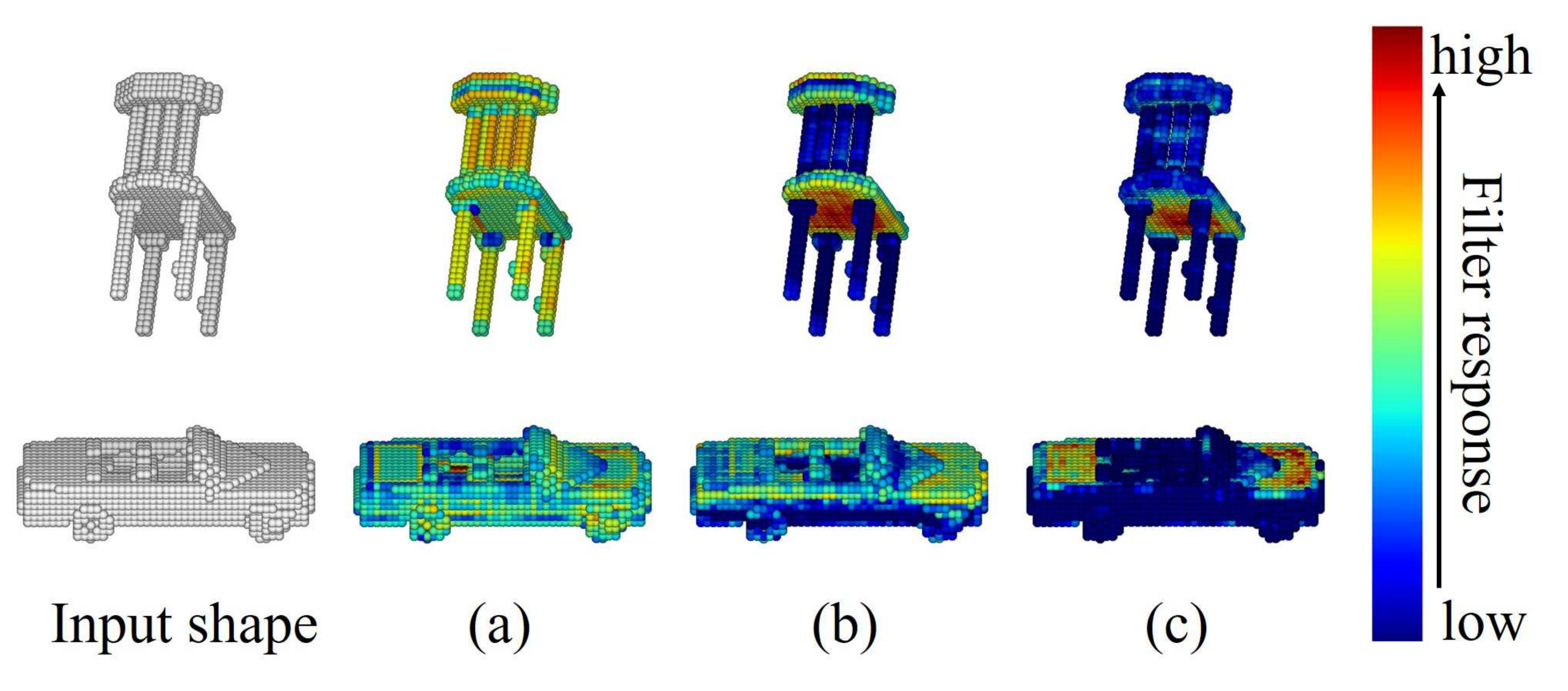}
	\caption{Visualization of filter responses from different ARB modules in Atrous 3DCNN. For each row, (a) is the response for the first ARB module; (b) for the second ARB module ; (c) for the third ARB module. The responses show filters in higher layer ARB could capture more discriminative local patterns.}
	\label{dilated_conv_fea}
\end{figure}

2) Atrous 3DCNN. Instead of large convolution strides and spatial pooling which could introduce information loss during feature extraction, dilated convolution is utilized to enlarge receptive field and preserve the spatial resolution simultaneously. In the atrous 3DCNN architecture, three Atrous Residual Blocks (ARBs) with dilated rates $2,3,4$ respectively are applied to extract features from different scales. Similar to U-Net, we call this part the feature encoding phase. After that, three residual blocks with skip layer connections to the encoding phase feature maps are used to aggregate different levels of features just like U-Net.

Intuitively, there should have been some improvement with the spatial resolution being preserved. However, we observe no better performance than U-Net 3DCNN. In order to investigate the reason, we visualize the filters by color-coding the responses to the input shape, just like what we do in CNN visualization (Section \ref{cnn_vis}). In Figure \ref{dilated_conv_fea}, (a) is the filter response from the first ARB with dilated rate 2. (b) and (c) are the filter responses from the second and third ARBs. As we can see, the filters in the first ARB (a) are activated in many regions across the input shape, which indicates that the filters are failed to capture discriminative features. Those less effective feature maps are then concatenated to higher layer feature maps by skip layer connection, decreasing the quality of the features. As discussed in Section \ref{3D_Atrous}, for the input data with sparse structure, it is highly possible that an atrous kernel could not cover local information due to the sparsity of both the input signal and the kernel weights. We also find that (b) and (c) demonstrate more discriminative filter responses (the activation distributed on certain patterns). Therefore, we assume stacked atrous convolutional layers could extract more robust features.

3) SDE+concat. To capture robust features from the sparse volumetric data, this network architecture uses three atrous residual blocks to form a Spatial Dense Extraction (SDE) unit. The encoding phase consists of two SDE units with dilated rates $[1,1,1]$ and $[1,3,5]$ respectively. The first SDE unit extracts local features, while the second SDE unit extracts features encoding long-range information. After that, the different levels of features are concatenated directly along the feature channel axis. Finally, the combined feature maps are fed to three stacked 3D convolutional layers with unit kernel size and unit stride to get the voxel-wise prediction. This architecture achieves average IoU of $84.34\%$ ($+0.37\%$ compared to baseline), demonstrating the effectiveness of our proposed SDE to extract features from sparse voxel data.

\begin{figure}[!t]
	\centering
	\includegraphics[width=1.0\linewidth]{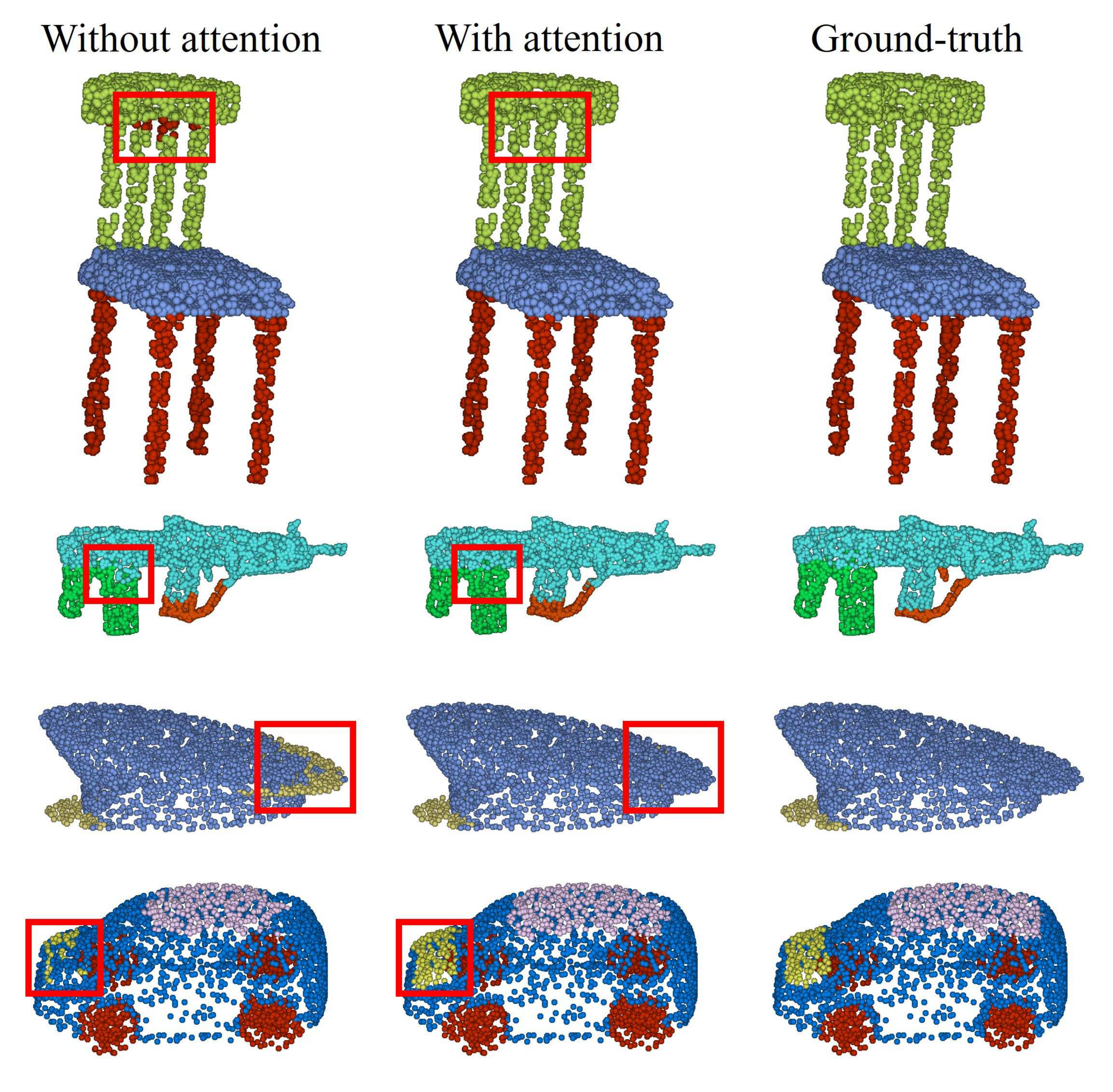}
	\caption{Our proposed attention feature aggregation (AFA) benefits part segmentation. There are several comparisons between segmentation results with and without AFA. For each object category, the first column is the direct concatenation result (SDE+concat); the second column is the attention feature aggregation result (SDE+AFA); and the third column is the ground truth segmentation. The comparison regions are highlighted by red boxes.}
	\label{compare_attention}
\end{figure}

4) SDE+AFA. Similar to SDE+concat network, this architecture uses SDE to extract different levels of features. In feature aggregation step, instead of direct concatenation, the Attention Feature Aggregation (AFA) is applied to combine the features extracted by different spatial dense extraction units. As multi-scale context is introduced, for a certain scale of thing, the features have different extent of discrimination. The attention mechanism could select the discriminative and effective features according to inputs. Specifically, the low-stage feature maps are re-weighted with the help of high-stage features before combination. Since the high-stage features usually capture high-level semantic regions, through attention feature aggregation, the low-stage feature maps activated by basic geometric patterns could be selected to better discriminate semantic parts. This architecture achieves average IoU of $86.24\%$ ($+1.90\%$ compared to SDE+concat).

In figure \ref{compare_attention}, several part segmentation results with and without the use of attention feature aggregation are presented. As we can see, with the help of AFA, the miss predicted chair back could be separated from the chair leg which shares similar low-level geometric patterns. Similar observations can be made for other shapes such as hat and car. That demonstrates the effectiveness of attention feature aggregation operation for semantic discriminative features extraction.

%
%

\begin{table}[t]  
	\small
	\renewcommand{\arraystretch}{1.2}
	\newcommand{\tabincell}[2]{\begin{tabular}{@{}#1@{}}#2\end{tabular}}
	\caption{Comparison between different network architectures}
	\label{tabel_2}
	
	\centering

	\begin{tabular}{c|p{25pt}||p{35pt}p{35pt}} 
		\hline
		Method &IoU  &\tabincell{l}{Precision\\ ($>3$ labels)} &\tabincell{l}{Recall\\ ($>3$ labels)} \\
		\hline
		U-Net 3DCNN  &83.97  &90.76 &92.92 \\
		Atrous 3DCNN &83.71  &91.14 &92.40 \\
		SDE+concat &84.34  &91.01 &93.01 \\
		SDE+AFA(2) &86.24  &92.12 &93.27 \\
		SDE+AFA(3) &\textbf{87.46}  &\textbf{92.98} &\textbf{93.72} \\
		\hline
	\end{tabular}	
\end{table}

\begin{figure}[!t]
	\centering
	\includegraphics[width=1.0\linewidth]{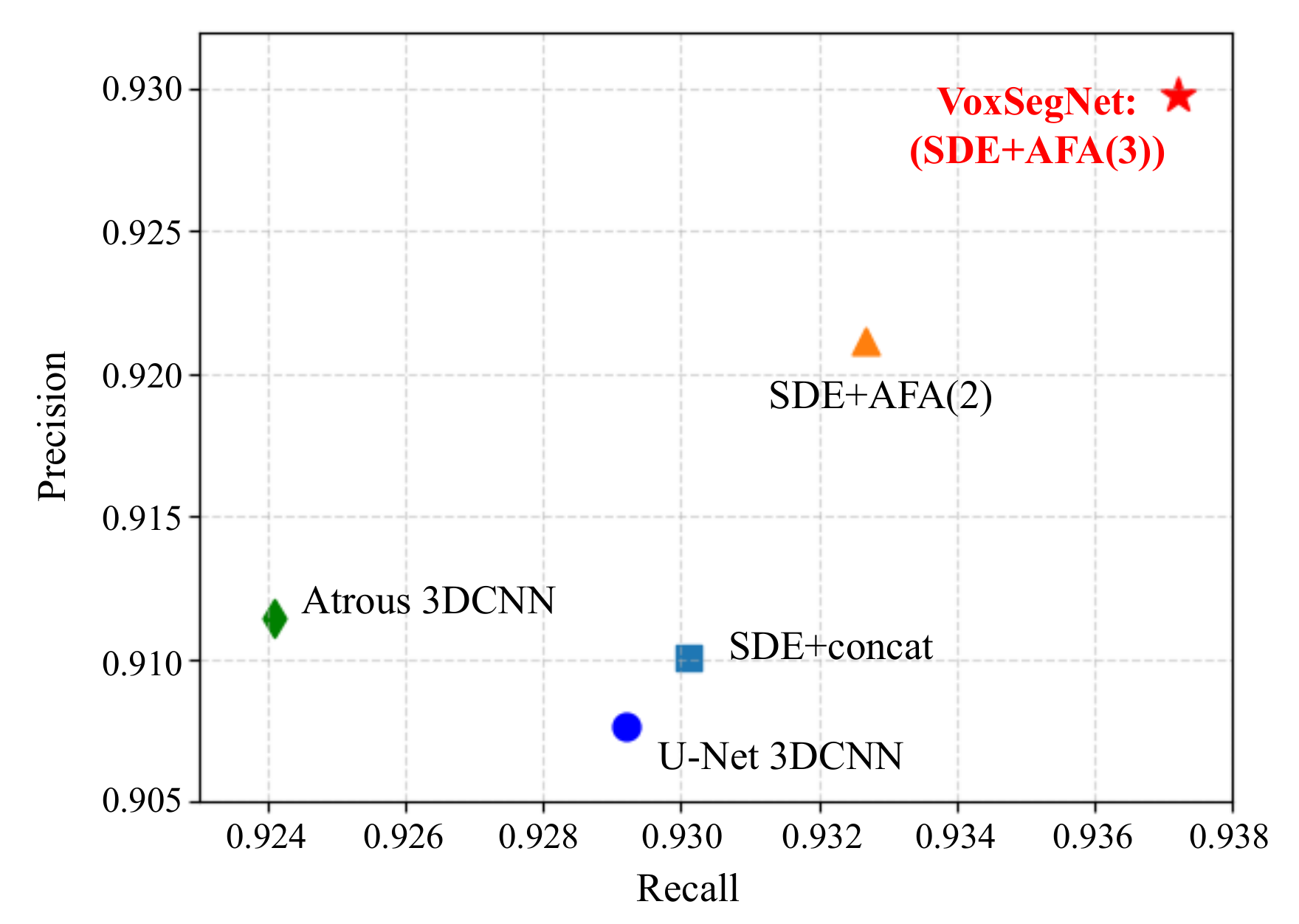}
	\caption{Visualized precision-recall comparison between different network architectures (on categories with more than 3 part labels).}
	\label{precision_recall}
\end{figure}

The performances of these network architectures are reported in table \ref{tabel_2}. In that table, besides the average IoU across all categories, we also compute the precision and recall values on a subset of object categories with more than 3 part labels (Airplane, Car, Chair, Lamp, and Motorbike). The corresponding precision-recall values are visualized in Figure \ref{precision_recall}. In summary, To evaluate the effect of our proposed modules, we add SDE and AFA to the baseline network gradually. Comparing baseline (U-Net) with the proposed networks, it is clear that leveraging SDE and AFA improves the performance by a large margin. The results show that each of our proposed modules can improve the performance of the corresponding network. We also find the network structure with three SDE units (applied in our VoxSegNet) outperforms that with two SDE units, utilizing more information from different scales. It also has a larger perceptive filed of 43 voxels with input spatial resolution of 48.

\subsection{Application: fine-grained shape clustering}

\begin{figure}[!t]
	\centering
	\includegraphics[width=1.0\linewidth]{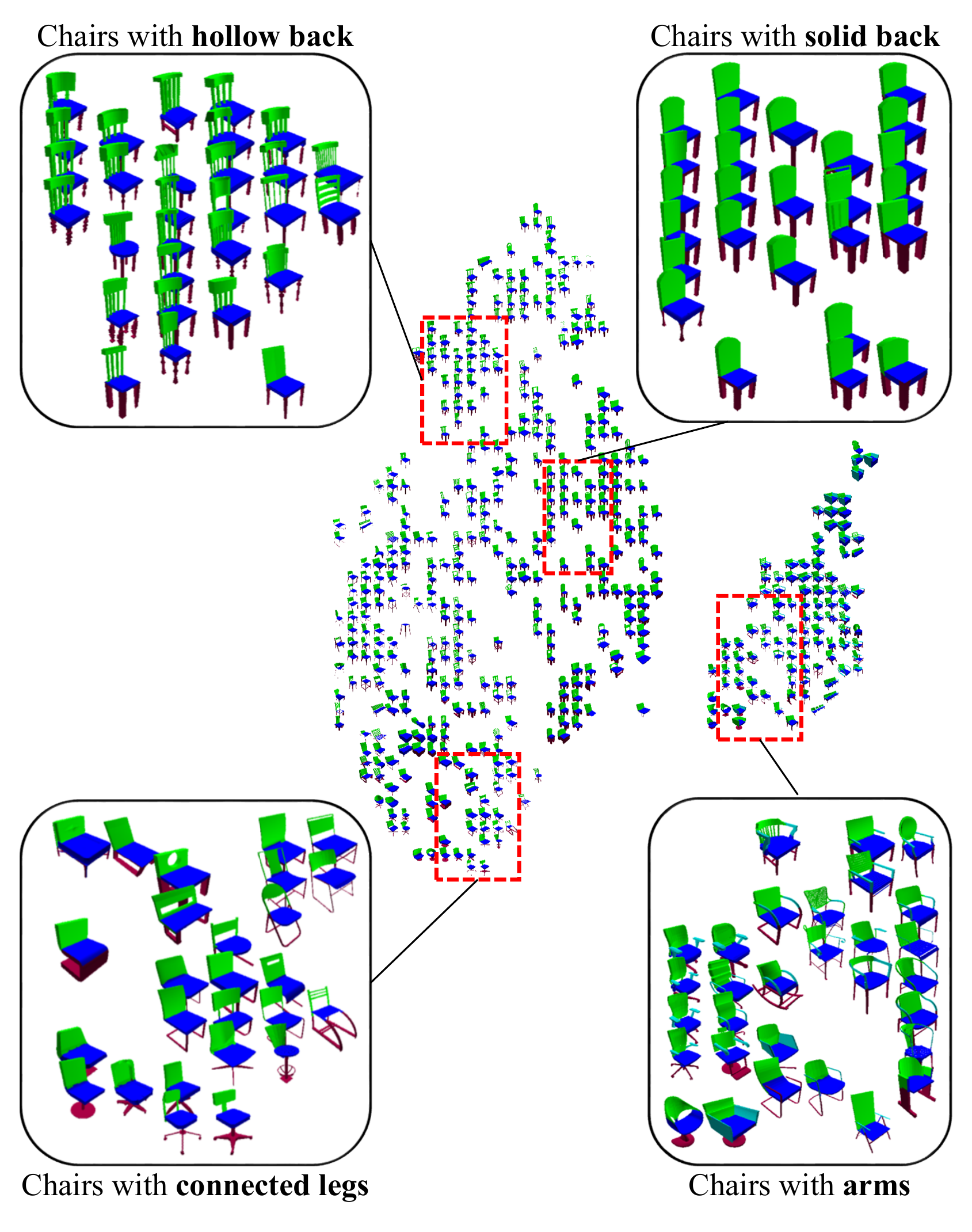}
	\caption{Fine-grained clustering results using the proposed part-based features. This is a t-SNE visualization. We can observe that the chairs in the same dashed box share similar geometric structure, which demonstrates the good description ability of part-based features. 
	}
	\label{fine_classification}
\end{figure}

With the emergence of large shape collections, the shapes within each category exhibit significant variation. For example, chair models from the Trimble 3D Warehouse could be further classified into sub-classes such as chairs-with-arms, swivel chairs, rocking chairs, etc. Fine-grained category information is important for shape understanding and could benefit exploration of the variability of a shape collection.

In this subsection, we show that with the help of semantic part information, fine-grained categories among a specific parent class could be investigated. In particular, given a collection of shapes from the same class (eg., chair), semantic part segmentation is conducted using our VoxSegNet. Then, we take the feature maps before $1\times1\times1$ convolutional layers as the shape descriptor and extract features per part. In detail, for each semantic part, the feature maps are multiplied element-wise by a mask (according to the part existence) and then averaged along spatial axes, generating a $64$ (number of channels) dimensional feature. The features from different parts are concatenated to form the part-based feature, which could describe a 3D model. Such features contain structural information, which could benefit fine-grained discrimination tasks. For example, in the chair category, the feature consists of four parts (back, seat, leg, and arm).

In Figure \ref{fine_classification}, we show a t-SNE visualization of the part-based features extracted from the chair category. As we can see, chairs with arms are separated from that without arms, and shapes with similar geometric structures are near each other in the feature space. Notice that the features are not specifically designed for object classification. In fact, semantic part segmentation features might be ambiguous for different detail geometric patterns in the same part category. Despite such drawback, the simple part-based features perform well in clustering the chairs according to shape structures.

\section{Conclution}
In this paper, we have presented a novel deep neural network for object part segmentation. Our method is motivated by extracting features better encoding detailed information under a limited resolution. Specifically, we introduce SDE to alleviate the detailed information loss caused by the sub-sampling procedure and the data sparsity. Moreover, we propose AFA to fuse the multi-scale information produced by SDEs through an attention mechanism. The experiment results on the large-scale shape part annotation dataset \cite{Yi16} and the ablation studies demonstrate the effectiveness of our proposed method.

\begin{figure}[!t]
	\centering
	\includegraphics[width=1.0\linewidth]{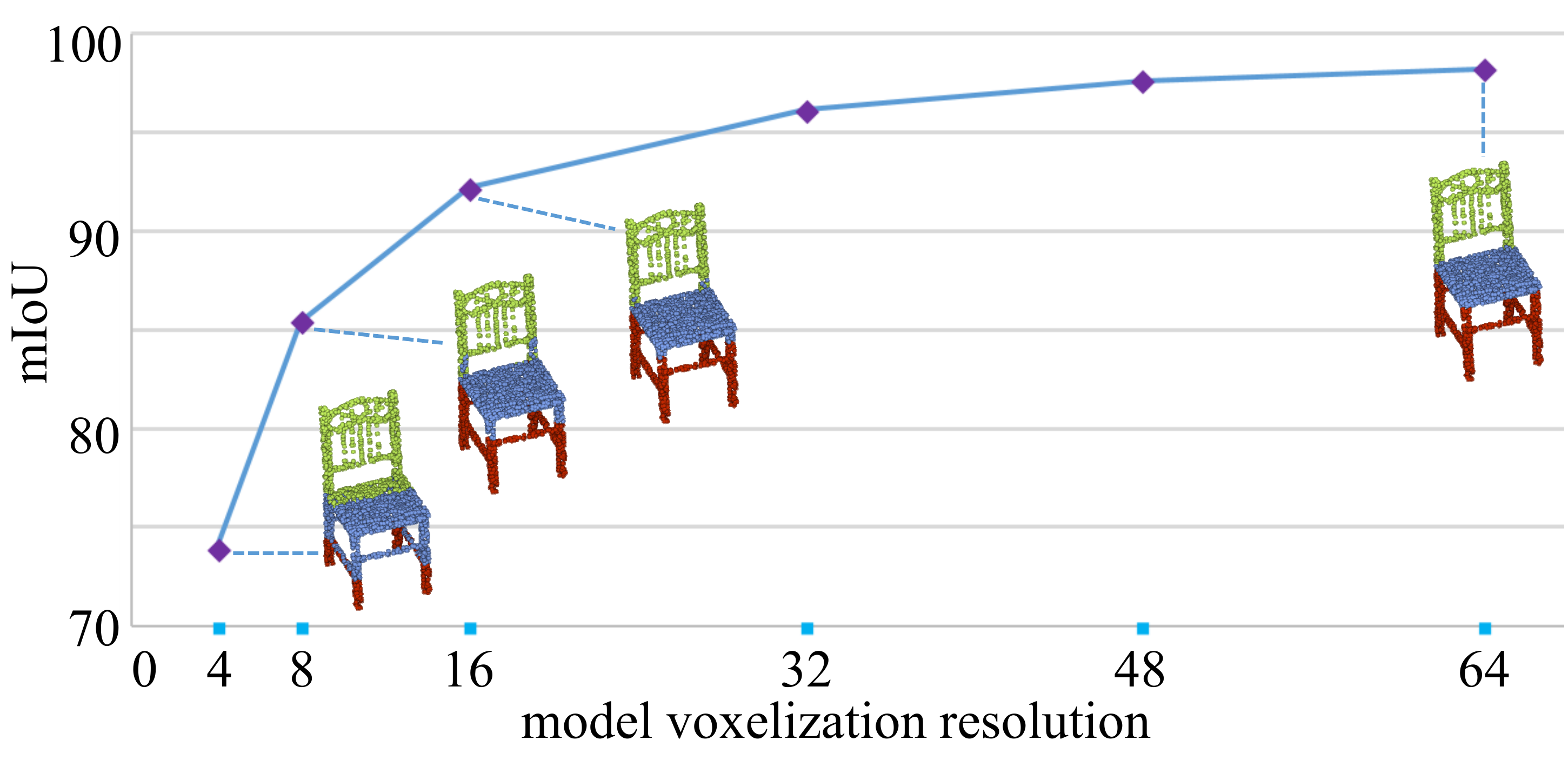}
	\caption{Finding the segmentation upper bound.
		We change the voxel resolution and evaluate the segmentation performance by projecting the voxel labels to the corresponding point clouds.
	}
	\label{quantization}
\end{figure}

\textbf{Future work.} In order to further improve the segmentation accuracy, one can follow the previous work \cite{Riegler2017} to use larger resolution volumetric data, or to extract finer features better describing the detailed structures as we did in this paper. We argue that the latter way possesses much potential. As shown in Figure \ref{quantization}, we performed an experiment to investigate the segmentation upper bound under different resolutions. Although a larger resolution results in a higher mIoU, it is harder to improve the upper bound as the resolution increases. In addition, there is still unignorable gap between the performance upper bound and the state-of-the-arts, indicating large room of improvement. Therefore, we believe that in the future, methods focusing on finer feature extraction from constrained resolution are worthy of researching.

\bibliographystyle{IEEEtran}
\bibliography{references}

\begin{IEEEbiography} [{\includegraphics[width=1in,height=1.25in,clip,keepaspectratio]{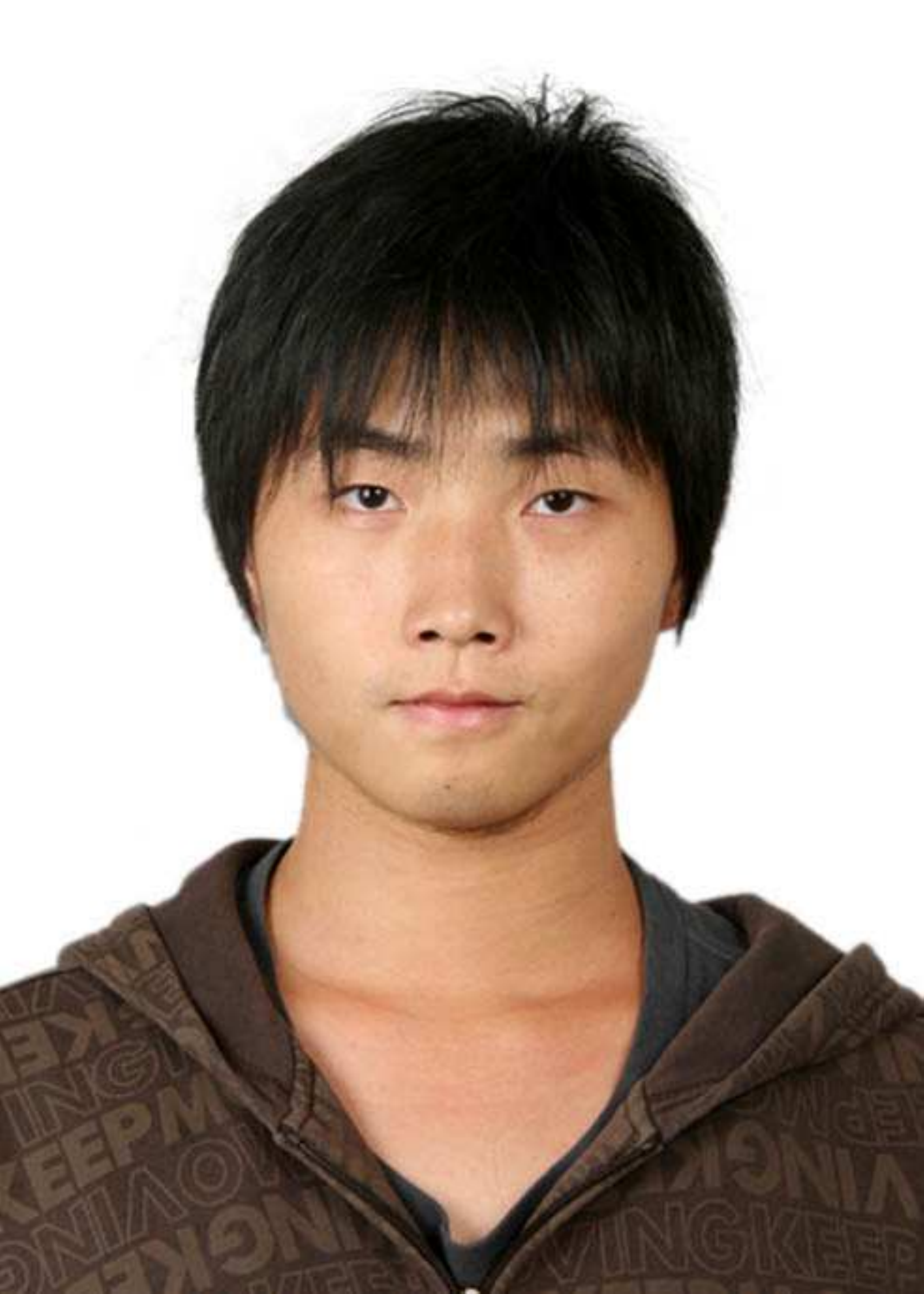}}]{Zongji Wang}
received the B.S. degree in School of Mathematics and Systems Science at Beihang University in 2014. He is currently pursuing the Ph.D degree with the State Key Laboratory of Virtual Reality Technology and System, School of Computer Science and Engineering, Beihang University. His research interests include computer vision and computer graphics.
\end{IEEEbiography}

\begin{IEEEbiography}[{\includegraphics[width=1in,height=1.25in,clip,keepaspectratio]{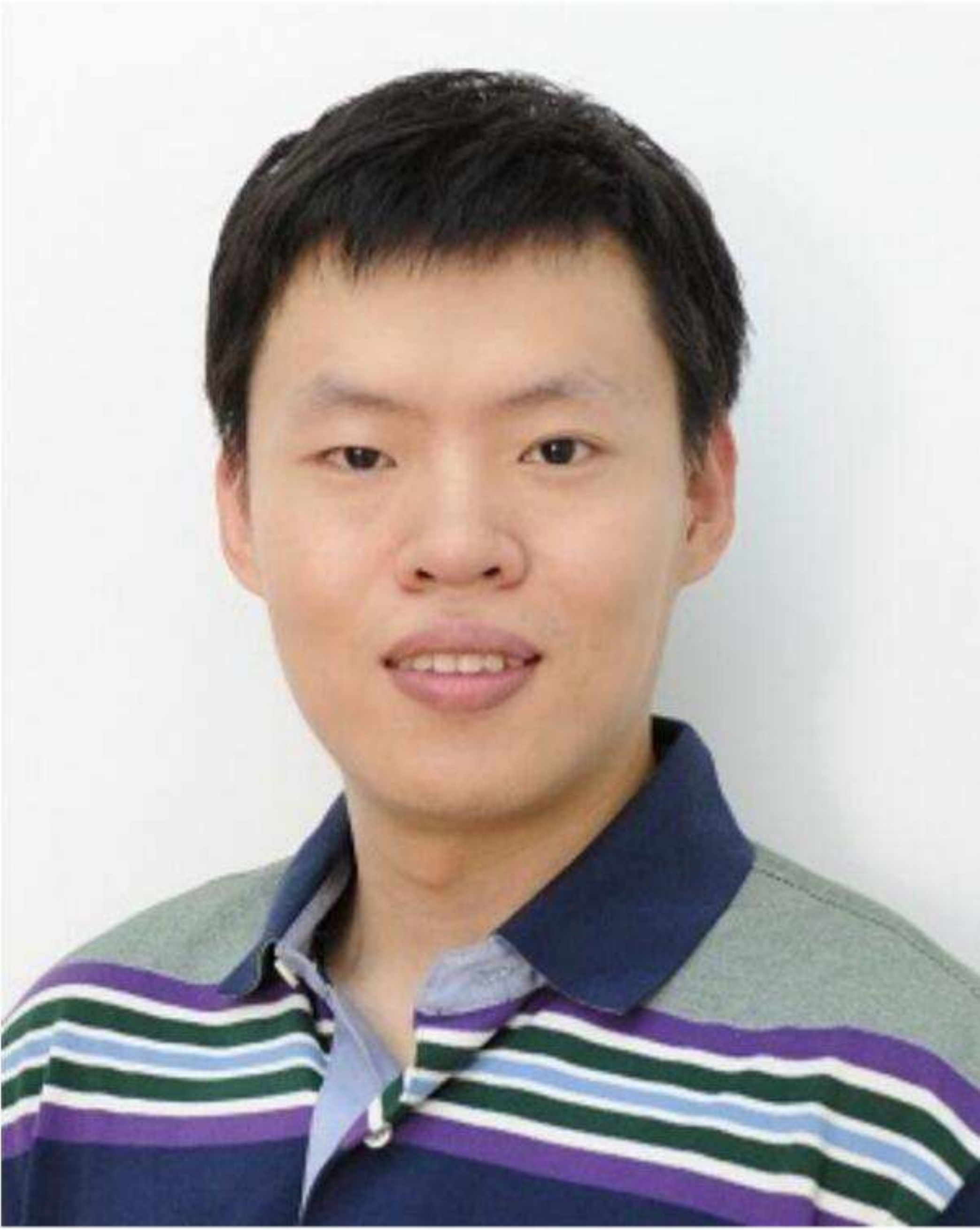}}]{Feng Lu}
received the B.S. and M.S. degrees
in automation from Tsinghua University, in 2007
and 2010, respectively, and the Ph.D. degree
in information science and technology from The
University of Tokyo, in 2013. He is currently
a Professor with the State Key Laboratory of
Virtual Reality Technology and Systems, School
of Computer Science and Engineering, Beihang
University. His research interests including 3D
shape recovery, reflectance analysis, and human
gaze analysis.
\end{IEEEbiography}

%
%




\end{document}